\documentclass[conference]{IEEEtran}
\IEEEoverridecommandlockouts
\usepackage{cite}
\usepackage{amsmath,amssymb,amsfonts}
\usepackage{algorithmic}
\usepackage{graphicx}
\usepackage{textcomp}
\usepackage{xcolor}
\usepackage[export]{adjustbox}
\def\BibTeX{{\rm B\kern-.05em{\sc i\kern-.025em b}\kern-.08em
    T\kern-.1667em\lower.7ex\hbox{E}\kern-.125emX}}
\usepackage[T1]{fontenc}
\usepackage{url}
\begin{document}

\title{Measuring Happiness Around the World Through Artificial Intelligence}

\author{
\IEEEauthorblockN{Rustem Ozakar}
\IEEEauthorblockA{\textit{Department of Computer Engineering} \\
\textit{Erzurum Technical University}\\
Erzurum, Turkey \\
rustem.ozakar@erzurum.edu.tr}
\and
\IEEEauthorblockN{Rafet Efe Gazanfer}
\IEEEauthorblockA{\textit{Department of Computer Engineering} \\
\textit{Erzurum Technical University}\\
Erzurum, Turkey \\
rafet.gazanfer27@erzurum.edu.tr}
\and
\IEEEauthorblockN{Y. Sinan Hanay}
\IEEEauthorblockA{\textit{Department of Computer Engineering} \\
\textit{Erzurum Technical University}\\
Erzurum, Turkey \\
sinan.hanay@erzurum.edu.tr}
}

\maketitle

\begin{abstract}
In this work, we analyze the happiness levels of countries using an unbiased emotion detector, artificial intelligence (AI). To date, researchers proposed many factors that may affect happiness such as wealth, health and safety. Even though these factors all seem relevant, there is no clear consensus between sociologists on how to interpret these, and the models to estimate the cost of these utilities include some assumptions. Researchers in social sciences have been working on determination of the happiness levels in society and exploration of the factors correlated with it through polls and different statistical methods. In our work, by using artificial intelligence, we introduce a different and relatively unbiased approach to this problem. By using AI, we make no assumption about what makes a person happy, and leave the decision to AI to detect the emotions from the faces of people collected from publicly available street footages. We analyzed the happiness levels in eight different cities around the world through available footage on the Internet and found out that there is no statistically significant difference between countries in terms of happiness.

\end{abstract}


\section{Introduction}

\par Emotions are a distinctive part of human nature. It is an essential element of life, and it contributes to our interactions with other people. Our intelligence attaches great importance to how we feel and distinguishes between different emotions. It is not possible to define or measure emotions in an exact manner, since they are abstract and subjective. However, social scientists have used different indicators about how one feels and what affects one's feelings. Thus, measuring happiness is possible to some extent. 

\par  Happiness is a multidimensional complex fact \cite{ram2010social}. It is a driving force and natural goal for most people. Researchers in social sciences have different methods on measuring happiness and variables that affect happiness. 

\par Analyzing happiness levels or life satisfaction of countries  attracted wide interest in social sciences and media. Generally, social sciences use polls to measure the level of happiness of society. Acquiring accurate results with happiness levels is a desired goal in different research areas. Machine learning can help with this problem by adding unbiased approach on measuring happiness and add an extra dimension to researchers in social sciences. In this mindset, our work examines the emotions from facial images collected from publicly available street footages of eight different cities around the world. Facial expressions from raw city footage are categorized into seven basic emotions; Anger, Sad, Neutral, Disgust, Surprise, Fear, Happy.

\par  Our approach is fundamentally different than that of indicators used by social scientists. It is relatively unbiased, because it relies on machine learning to determine the emotions of people. Machine learning has been a powerful tool for solving problems in various domains such as medicine, economics and robotics. It can accurately detect emotions by analyzing the facial images.

\par The organization of the paper is as follows; related works in this area are discussed in Section II. Section III describes the proposed methods in detail. Results are given in Section IV and Section V concludes the paper.

\section{Related Work}

%Understanding the factors that affect happiness have been %studied extensively. Some research analyzed the effect of %wealth on happiness []
Understanding the factors that affect happiness have been studied extensively. Veenhoven, Ruut and Ehrhardt \cite{veenhoven1995cross} investigated if the data available on happiness are in accordance with the three theories of happiness; comparison, folklore and livability. In a previous study, the researchers concluded that happiness is adversely affected for the people when their neighboring countries becomes wealthier \cite{becchetti2013beyond}. In another study \cite{ram2010social}, it is investigated whether social capital has a significant effect on happiness. In \cite{easterlin2010happiness}, researchers updated the Easterlin Paradox, the idea that happiness is proportional to income, is valid for only developed countries. They concluded that this paradox also holds for less developed countries. In \cite{sachs2018world}, the effects of different variables on the happiness (migration, health issues, income, etc.) are investigated on both national and global scale. HSBC's expat survey \cite{hsbcExpat} measured the satisfaction of expats in different aspects for various countries.

\par There are varying approaches on the subject in literature for emotion recognition ranging from examining brain signals to hybrid approaches like analyzing audio and text input together with videos. 
Machine learning algorithms such as  Support Vector Machines (SVM) \cite{svm92}, Neural Networks and K-nearest Neighbors (k-NN) can be used to predict emotions from face images.

%Common methods that are used for emotion detection in computer science can be listed as follows; Active Shape Model, Fuzzy Logic, Gabor Filters, Gaussian Mixture Models (GMM), Histogram of Oriented Gradients (HOG), Hidden Markov Model (HMM), Kalman Filter, Local Binary Pattern, Bayesian Classifiers, Deep Neural Networks, Principal Component Analysis (PCA), Linear Discriminant Analysis (LDA), Particle Swarm Optimization (PSO), Support Vector Machines (SVM) \cite{svm92}, Viola-Jones algorithm, Optical Flow Analysis.

\par Some of the notable works in this field are as follows; Ekman and Friesen \cite{ekman1980facial} developed a system which is called FACS (Facial Action Coding System), where basic universal facial expressions are represented as combinations of different action units. Black and Yacoob \cite{black1995tracking} used polynomials to represent optical flow and extract the motion of the facial features. After that, they used a rule-based classifier to classify basic emotions. Yacoob and Davis \cite{yacoob1996recognizing} used optical flow of the detected facial features within rectangular regions with rule-based classifier for recognition. Essa and Pentland \cite{essa1997coding} transformed face images into mesh models, then calculated the optical flow from them to recognize facial expressions. Donato et al. \cite{donato1999classifying} compared different techniques to extract and classify facial expressions. Cowie et al. \cite{cowie2001emotion} published a survey about emotion recognition containing detailed information about both audio and visual approaches. Cohen et al. \cite{cohen2003facial} in their work used Naive Bayes Classifiers with a Cauchy distribution together with HMM for facial expression recognition. Ioannou et al. \cite{ioannou2005emotion} used a combination of SVM, morphological operators, neural networks and neuro-fuzzy system to classify facial expressions. Gunes and Piccardi \cite{gunes2007bi} used face and hand gestures in combination to determine emotional expression. Mansoorizadeh and Charkari \cite{mansoorizadeh2010multimodal} introduced a hybrid feature space from extracted features of speech and video and performed emotion recognition. Dahmane and Meunier \cite{dahmane2011emotion} used HOG and SVM to classify emotions from videos containing face. Halder et al. \cite{halder2013general} used different combinations of fuzzy sets to classify emotions from faces. Poria et al. \cite{poria2016convolutional} used Convolutional (CNN) and Recurrent Neural Networks (RNN) with Multiple Kernel Learning to analyze sentiment from audio, video and text. Zhang et al. \cite{zhang2016facial} introduced a fusion approach using a Part-based Hierarchical Bidirectional RNN (PHRNN) and a Multi-Signal CNN Network (MSCNN) on images to recognize emotions. Jain et al. \cite{jain2018hybrid} also proposed a CNN-RNN hybrid model on face images for recognition. Hossein and Muhammad \cite{hossain2019emotion} worked on audio and image inputs together using CNN and SVM for emotion recognition.\\

\section{Methodology}\label{methodology}

In our work, we present an unbiased emotion detector through machine learning using publicly available videos of the crowded streets in various countries. We aimed to gather unbiased data from different cities around the world to analyze the psychology of the general population. Footages from different cities are converted to frames to be used as dataset. All methods mentioned here are performed with Python language.
\par To detect facial emotion recognition, first we need to detect faces in each frame in the dataset. For this purpose, we used Adam Geitgey's \cite{geitgey2017face} face recognition model. With this model, it is possible to locate face in the frame and crop the face region for further processing.
\par For each city, we traverse through all frames and detected faces are cropped, turned into grayscale and resized to 48x48 pixels using OpenCV library. Then, these images are sent to pre-trained deep learning model of Priya Dwivedi \cite{dwivedi2019} as inputs. Classification result belongs to one of these emotion categories; Anger, Sad, Neutral, Disgust, Surprise, Fear, Happy. Finally, classification results are exported to a CSV file categorized by city, emotion and value.

\subsection{Dataset}

\par Our dataset contains eight cities, Barcelona, Copenhagen, Istanbul, Kiev, London, New York, Paris and Tokyo. Footages were taken from various publicly available videos online, containing crowded street scenes dated between July 2013 - October 2020. Each city contains 300 detected 48x48 pixel face images in grayscale, thus complete dataset has 2400 images. It is made sure that dataset doesn't have any mistaken image for face, or, any sunglasses or regular glasses present.\\

\subsection{Face Recognition}

For face recognition, we used Adam Geitgey's \cite{geitgey2017face} CNN based face recognition library. It is actually a Python adaptation of Dlib's \cite{king2009dlib} face recognition algorithm, a popular C++ library that consists of various machine learning algorithms. CNN method can detect faces from different angles, unlike the HOG based version in the Dlib library.\\

\subsection{Emotion Recognition}

As previously mentioned, we used Priya Dwiedi's model \cite{dwivedi2019} for emotion recognition. They trained their model with The Facial Expression Recognition 2013 (FER-2013) database which is a work of Pierre-Luc Carrier and Aaron Courville \cite{goodfellow2013challenges}. This database consists of 28,709 face images for training in 48x48 pixel grayscale format.
\par A graphic representation of the CNN model prepared with visualkeras \cite{visualkeras} can be seen in the Fig. 1. Activation functions used in layers are ReLU, except the last layer, which is softmax.\\

\begin{figure*}[htbp]
    \centering
    \includegraphics[width=0.75\textwidth]{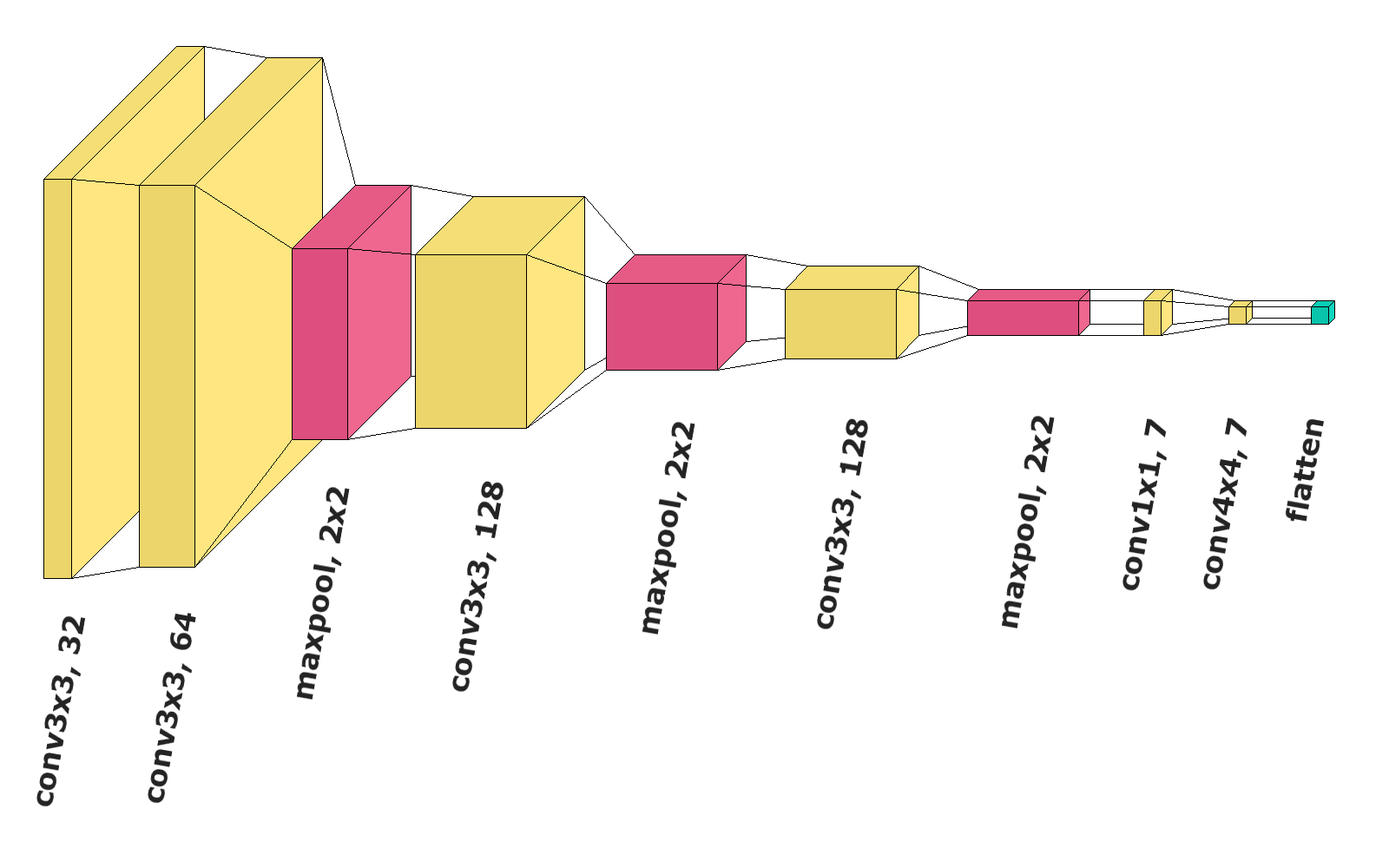}
    \caption{CNN architecture for emotion recognition}
    \label{fig:my_label}
\end{figure*}

\section{Experimental Results}

\par Our results showed that most common emotions detected with the algorithm were; Surprise, Fear, Happy and Anger. Sad and Neutral emotions were not detected in any of the cities. In Table-I detailed results containing all emotions are given. We notice that "surprise" emotion is very common. One explanation can be due to the recording nature of video (i.e. a youtuber recording the video) people might be surprised to notice a camera.

\par In Fig. 2, 95\% confidence intervals for happiness proportion by cities are shown. The intervals show that there is no statistically significant difference between cities in terms of proportion of happy people. There can be several explanations for that. First, even though we collected street footage, the recordings used were not taken in an obscure manner and this might have biased emotions. Another explanation could be that the proportion of happy people in street is indeed similar across the world. Results are discussed in Conclusion section.

\begin{table}[htbp]
\caption{Emotion results by city}
\begin{center}
\begin{adjustbox}{max width=9.0cm}
\begin{tabular}{|c|c|c|c|c|c|}
\hline
\textbf{} & \textbf{Anger} & \textbf{Disgust}  & \textbf{Surprise} &\textbf{Fear} & \textbf{Happy}\\
\hline
Barcelona & 2 & 0 & 90 & 203 & 5 \\
\hline
Istanbul & 1 & 1 & 64 & 227 & 7 \\
\hline
Kiev & 2 & 0 & 79 & 215 & 4 \\
\hline
London & 1 & 0 & 115 & 182 & 2 \\
\hline
New York & 1 & 0 & 61 & 230 & 8 \\
\hline
Paris & 2 & 0 & 31 & 267 & 0 \\
\hline
Tokyo & 3 & 0 & 32 & 260 & 5 \\
\hline
Copenhagen & 1 & 0 & 69 & 227 & 3 \\
\hline
\end{tabular}
\end{adjustbox}
\end{center}
\end{table}

\begin{figure}[htbp]
    \includegraphics[width=0.48\textwidth, center]{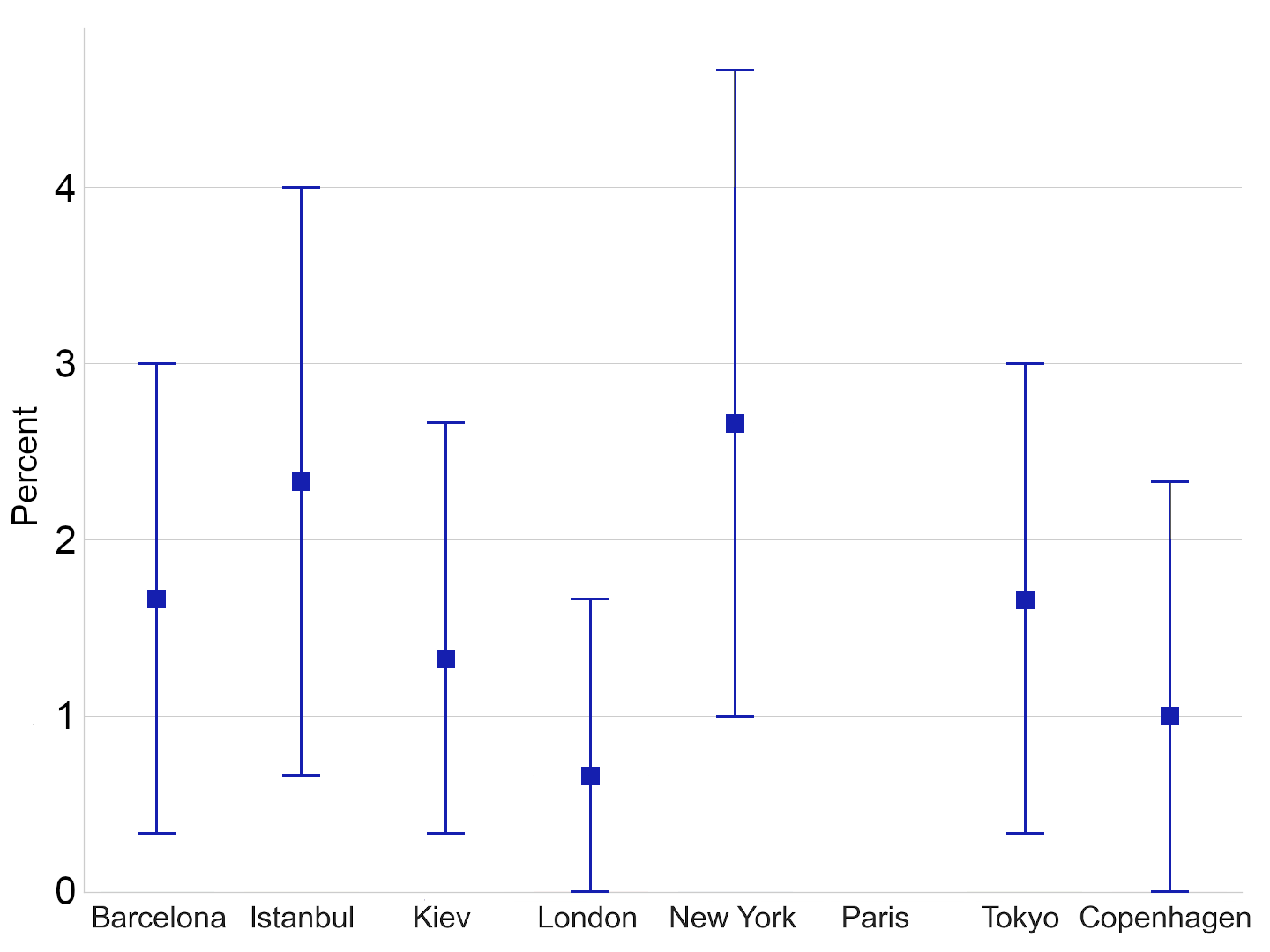}
    \caption{Happiness distribution by city}
    \label{fig:my_label}
\end{figure}

%block comment below, 
\iffalse
    \begin{table}[htbp]
    \caption{Happiness comparison by city}
    \begin{center}
    \begin{tabular}{|c|c|}
    \hline
    \textbf{City} & \textbf{Happiness \%} \\
    \hline
    New York & 2.66 \\
    \hline
    Istanbul & 2.33 \\
    \hline
    Barcelona & 1.66 \\
    \hline
    Tokyo & 1.66 \\
    \hline
    Kiev & 1.33\\
    \hline
    Copenhagen & 1.00 \\
    \hline
    London & 0.66 \\
    \hline
    Paris & 0\\
    \hline
    \end{tabular}
    \end{center}
    \end{table}
\fi

\section{Conclusion}

\par In our work, we proposed the use of public footage as a resource to determine happiness levels in society. We collected footage from various cities around the world together, then used artificial intelligence to recognize emotions from facial expressions.

\par For the reason why surprise and fear are prevalent in results, we think that it could be due to sun light directly hitting people outdoors, thus making them grimace, or maybe they were in a rush to catch up something, or maybe they were surprised when they noticed the camera recording. Another reason may be different races have different facial structures which can give a way to unintended classifications. It could be due to the neural network architecture, or the learning method altogether, it may need more training with different datasets or update in architecture.

\par For a future direction, further work needed in creating a more specialized dataset for the purposed idea and different machine learning algorithms needs to be explored in terms of performance and accuracy. 

\bibliography{references}
\bibliographystyle{IEEEtran}
%\nocite{*}
\end{document}